\documentclass{article}
\usepackage{iclr2027_conference,times}

\usepackage{amsmath,amsfonts,bm}

\def\eqref#1{equation~\ref{#1}}
\def\1{\bm{1}}

\DeclareMathAlphabet{\mathsfit}{\encodingdefault}{\sfdefault}{m}{sl}
\SetMathAlphabet{\mathsfit}{bold}{\encodingdefault}{\sfdefault}{bx}{n}

\usepackage{amsmath,amssymb}
\usepackage{booktabs}
\usepackage{longtable}
\usepackage{multirow}
\usepackage{array}
\usepackage{graphicx}
\usepackage{xcolor}
\usepackage{colortbl}
\usepackage{float}
\usepackage{flafter}
\usepackage{placeins}
\usepackage{capt-of}
\usepackage{wrapfig}
\usepackage{needspace}
\usepackage{url}
\usepackage{hyperref}
\hypersetup{hidelinks}
\definecolor{fabricblue}{HTML}{4C6A92}
\definecolor{fabricpale}{HTML}{EEF3F8}
\definecolor{fabricmint}{HTML}{EFF6F3}
\definecolor{fabricrule}{HTML}{35506E}

\makeatletter
\newcommand{\compactwrapcaption}{%
  \long\def\@makecaption##1##2{%
    {\normalfont\normalsize\raggedright ##1: ##2\par}%
    \vskip4pt}}
\renewcommand{\@maketitle}{%
  \vbox{\hsize\textwidth\centering
    {\LARGE\scshape\@title\par}\vskip14pt
    {\normalfont\normalsize\@author\par}\vskip16pt}}
\makeatother

\title{VLA-Fabric: Learning Interaction\\for Composable Multi-Arm VLA Agents}

\author{%
  \textbf{Chenglong Zhang\textsuperscript{1}}\qquad
  \textbf{Shuaijun Liu\textsuperscript{1}}\qquad
  \textbf{Feiyang You\textsuperscript{1}}\\[3pt]
  \textbf{Shuyang Hao\textsuperscript{1}}\qquad
  \textbf{Chengyu Wu\textsuperscript{1}}\qquad
  \textbf{Ningxin Su\textsuperscript{1,*}}\\[7pt]
  {\small\textsuperscript{1}The Hong Kong University of Science and Technology (Guangzhou)}\\[3pt]
  {\small\textsuperscript{*}Corresponding author:
    \href{mailto:ningxinsu@hkust-gz.edu.cn}{ningxinsu@hkust-gz.edu.cn}}\\[4pt]
  {\small Project Website:
    \href{https://nebulis-lab.com/VLA-Fabric/}{\textcolor{fabricblue}{nebulis-lab.com/VLA-Fabric}}}}

\iclrfinalcopy
\hypersetup{pdftitle={VLA-Fabric: Learning Interaction for Composable Multi-Arm VLA Agents}}
\hypersetup{pdfauthor={Chenglong Zhang, Shuaijun Liu, Feiyang You, Shuyang Hao, Chengyu Wu, Ningxin Su}}

\begin{document}
\maketitle
\lhead{Preprint}
\raggedbottom

\begin{abstract}
Vision-language-action (VLA) models are rapidly expanding what individual
robots can accomplish, yet many real-world manipulation settings require
multiple arms to operate as a coordinated system. Existing approaches organize
multi-arm coordination in different ways, including centralized joint policies,
high-level task or role assignment, and locally executed policies that react
to teammates. However, how complete action-producing VLAs should explicitly
interact with one another during action generation remains largely unexplored.
We introduce VLA-Fabric, which organizes one complete VLA per arm while
preserving local perception and action ownership, and enables coordination
through learned interaction across policy boundaries. We first systematically
study where such interaction should occur, how peer information should be
aggregated, and how the interaction should be learned, revealing complementary
coordination functions spanning shared context, peer-specific latent
information, and action-stage interaction. We then re-instantiate these
functions in a structurally different VLA architecture, showing that the
coordination principle is not tied to a particular tensor layout or backbone
implementation. Finally, we compose teams of three and four arms
and evaluate them in simulation and physical experiments.
On all four physical tasks, Full interaction improves success over
independently trained local policies by 16--40 percentage points.
Together, these results suggest that explicit interaction among complete VLA
policies offers a scalable path toward networked embodied intelligence.
\end{abstract}

\section{Introduction}
\label{sec:introduction}

Vision-language-action (VLA) models are advancing the capabilities of individual
robots by connecting visual and language understanding with learned action
generation \citep{rt2,openvla,pi05}. Many important robotic applications,
however, require several manipulators to contribute to a common physical
objective. Cooperative manufacturing, multi-part assembly, shared-load handling,
and flexible robotic workcells motivate systems in which multiple arms work
together \citep{cooperative_topp,apexmr,fabrica}. As pretrained VLAs become more
capable, their progress creates an opportunity beyond improving a single robot:
organizing multiple action-producing policies into a collaborative system.
The ability of each constituent model to perceive and act does not, by itself,
specify how those models should work together.

Recent multi-arm VLAs pursue this goal through different organizations.
RDT-1B, ET-VLA, TwinVLA, and Dexora place cross-arm dependencies within a joint
policy \citep{rdt1b,etvla,twinvla,dexora}.
Other approaches coordinate at different levels: MA-VLA assigns
atomic behaviors to individual arms \citep{mavla}, while CHORUS demonstrates
implicit decentralized coordination through locally executed policies that
observe and react to teammates without runtime messages \citep{chorus}.
These approaches establish effective collaboration, with coordination inside a
joint policy, above the action policy through behavior assignment, or indirectly
through observations of teammates. This raises an organizational challenge as
the number and roles of manipulators grow. How can a multi-arm system remain
modular at the level of individual manipulators while still supporting tightly
coupled collaboration?

A natural way to obtain such modularity is to organize the system around
complete arm-local policies rather than a growing joint policy. We therefore
treat each manipulator and its complete VLA as a self-contained execution unit
that perceives its local context and produces its own actions.
Local information can remain with the policy that uses it,
and larger teams can be formed by composing additional execution units.
Crucially, \emph{local does not mean isolated}. Policies participating in the
same physical task still need to adapt their actions to one another. The key
challenge is to connect these local policies through explicit interaction
without giving up their local action ownership.

\begin{figure}[htbp]
\centering
\IfFileExists{figures/motivation_roadmap_asset.pdf}{%
  \includegraphics[width=\linewidth,height=0.51905\linewidth,keepaspectratio]{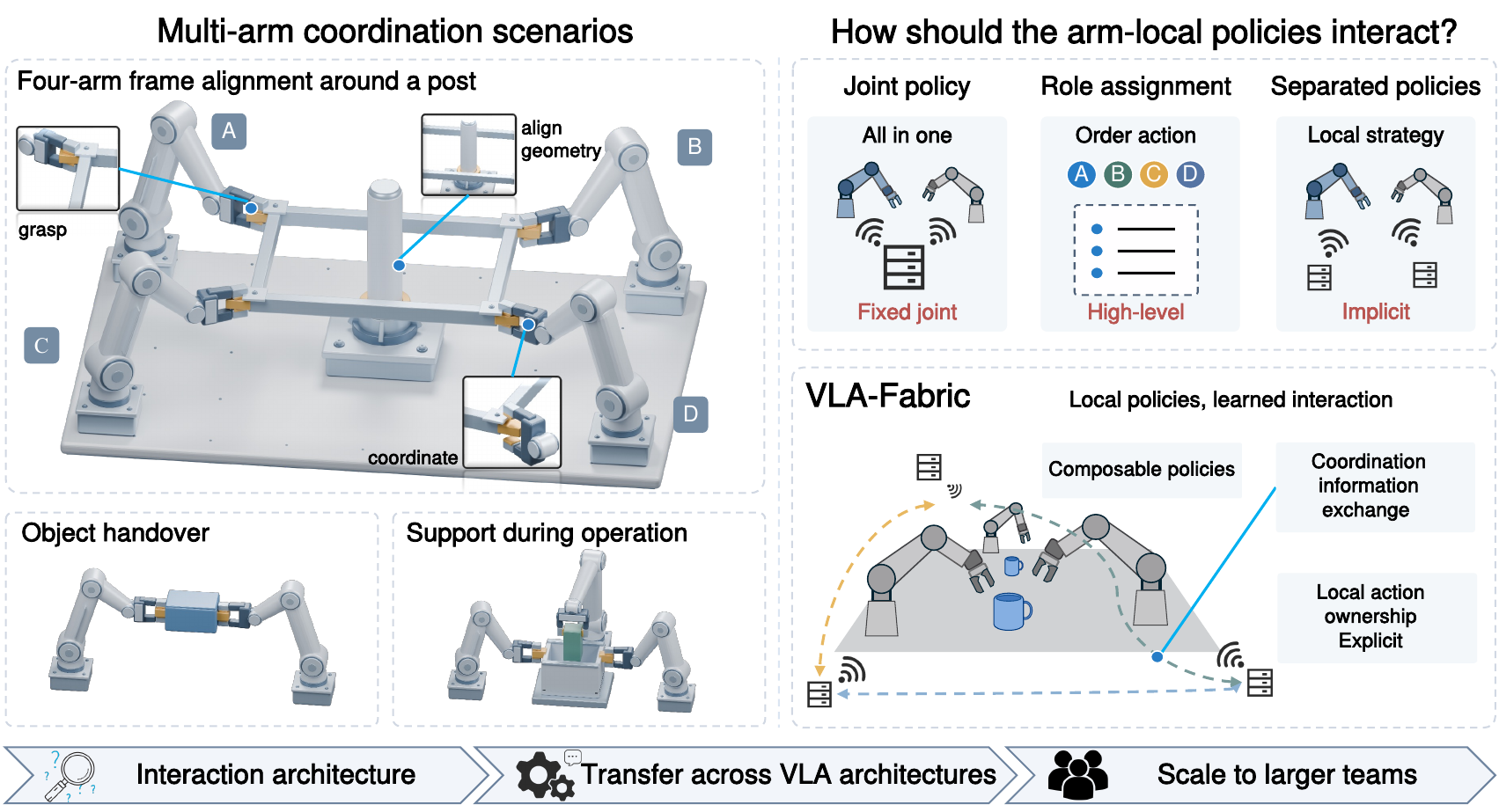}%
}{%
  \includegraphics[width=\linewidth]{figures/motivation_roadmap.png}%
}
\caption{From multi-arm coordination to VLA-Fabric. Complete arm-local VLAs
collaborate through learned interaction; we study interaction discovery,
cross-architecture transfer, and larger-team scaling.}
\label{fig:motivation}
\end{figure}

To realize this local-but-connected organization, we introduce
VLA-Fabric\footnote{Project website and code:
\url{https://nebulis-lab.com/VLA-Fabric/}.},
a multi-VLA coordination architecture connecting complete arm-local policies
through learned internal representation interaction
(Figure~\ref{fig:motivation}). Each VLA retains local perception and action
generation while selected peer representations condition its computation
through explicit policy interaction. VLA-Fabric thus makes coordination
among complete policies an explicit learning and architecture problem instead
of absorbing all cross-arm dependencies into a joint policy.

First, we establish an effective VLA-Fabric interaction architecture through a
broad controlled study. Starting from a decomposable bimanual VLA with
pretrained single-arm streams \citep{twinvla}, we separate its original cross-arm
dependencies. We systematically explore a combinatorial design space covering
interaction entry points, aggregation operators, and fine-tuning strategies.
The study shows that exposing more peer information does not by itself ensure
coordination; interaction structure and adaptation determine how effectively
local policies use one another's information. This exploration identifies
complementary interaction at shared-context, peer-specific intermediate, and
action-generation stages.

Second, we demonstrate that the same VLA-Fabric interaction paradigm can be
re-instantiated in substantially different VLA architectures. We use $\pi_{0.5}$
\citep{pi05} as a second realization, whose multimodal processing and action
generation differ markedly from the Eagle2 VLM-based SingleVLA used in our
initial study. Accordingly, the same coordination functions are realized at
architecture-appropriate computational boundaries while preserving complete
local policies with explicit policy interaction. The resulting
$\pi_{0.5}$-Fabric system enables multiple complete $\pi_{0.5}$ policies to
interact and achieves strong performance across several
bimanual tasks. Its stronger results suggest that advances in the constituent
VLA may translate into stronger collaborative systems. This observation is
consistent with the value of pretrained single-arm capabilities demonstrated
by TwinVLA \citep{twinvla}.

Third, we establish the extensibility of VLA-Fabric from bimanual coordination
to larger teams of complete VLA policies. Additional manipulators are
incorporated by composing additional local policies and extending their
interaction. We instantiate three- and four-arm VLA-Fabric teams and evaluate
them on simulation tasks with different coordination demands.
We further validate the same organization on four physical
three- and four-arm tasks, where Full interaction consistently improves over
independently trained local policies.
Our contributions are:

\noindent\textbf{VLA-Fabric interaction architecture.}
We introduce VLA-Fabric, which organizes complete arm-local VLAs as the execution
units of multi-arm collaboration and connects them through learned
representation interaction. A broad controlled architecture study establishes
effective interaction structure, aggregation, and adaptation for coordinated
action generation.

\noindent\textbf{Cross-architecture transfer of VLA-Fabric.}
We re-instantiate the same interaction paradigm in Eagle2 VLM-based SingleVLA
and $\pi_{0.5}$ despite their substantially different policy computations.
The resulting systems demonstrate a common coordination organization across
distinct VLA architectures.

\noindent\textbf{Composable multi-arm scalability.}
We scale VLA-Fabric from two complete VLA policies to teams of three and four
while preserving local action generation and explicit policy interaction.
We validate the resulting organization across simulation and physical
multi-arm tasks with differing coordination demands.

\section{Related Work}
\label{sec:related}

\paragraph{Cooperative physical manipulation.}
Multi-arm robotics has studied coordinated transport, cooperative assembly,
and learned dual-arm manipulation under shared physical constraints
\citep{cooperative_topp,apexmr,fabrica}. These settings require compatible
arm motions under a shared load or coordinated holding, alignment, and
placement during assembly. Planning and control address the resulting task
and motion dependencies. Our work studies the complementary policy-level
problem of organizing complete learned VLAs into a collaborative system.

\paragraph{Multi-arm VLA coordination.}
Multi-arm VLAs organize coordination through several mechanisms.
RDT-1B, ET-VLA, TwinVLA, and Dexora resolve cross-arm dependencies within
joint bimanual policies \citep{rdt1b,etvla,twinvla,dexora}.
MA-VLA uses high-level behavior assignment \citep{mavla}, while CHORUS
coordinates locally executed policies through teammate observations
without inference-time communication \citep{chorus}.
VLA-Fabric studies explicit learned interaction between complete arm-local
VLA policies during action generation.

\paragraph{Learned interaction among embodied policies.}
LatentToM combines ego and consensus representations in decentralized
diffusion policies and supports explicit consensus exchange for cooperative
manipulation \citep{latenttom}. Its no-message variant instead infers teammate
behavior from local observations.
CommNet, DIAL, and TarMAC establish broader foundations for learning
communication jointly with agent policies \citep{commnet,dial,tarmac}.
Here, each agent is a complete pretrained VLA containing perception,
multimodal reasoning, and action generation. We study which internal
coordination functions should connect these policies and whether the resulting
interaction organization transfers across substantially different VLA
architectures.

\section{VLA-Fabric: Learning Interaction among Local VLA Agents}
\label{sec:interaction}

\subsection{Local VLA Agents and the Interaction Problem}
\label{sec:problem}

VLA-Fabric organizes multi-arm manipulation around complete local execution
units, termed \emph{VLA agents}. Each agent runs a complete VLA policy and
owns one arm's perception-action loop, including its policy parameters,
inference state, private observations, proprioception, and action output.
Selected peer representations condition local action generation through
learned interaction. We use \emph{policy} for the learned mapping
$\pi_{\theta_i}$ and \emph{agent} for its execution with local ownership and
team interaction relations. Task-level context, such as an instruction or
shared scene observation, may be available to multiple agents without
changing their local action ownership.

For a team of $N$ agents, agent $i$ predicts an $H$-step action chunk at
planning round $t$:
\begin{equation}
a_i^{t:t+H-1}
=\pi_{\theta_i}(o_i^t,q_i^t,\ell,r_i;\mathcal M_i^t).
\label{eq:local_policy}
\end{equation}
Here $o_i^t$ and $q_i^t$ are the agent's visual and proprioceptive inputs,
$\ell$ is the task instruction, and $r_i$ is an optional role descriptor.
The context $\mathcal M_i^t$ contains peer information received during this
round of action generation. Independently trained policies keep
$\mathcal M_i^t=\varnothing$ throughout fine-tuning and inference.

The central design problem is the construction of $\mathcal M_i^t$ without
replacing the local action-generating policies. This requires specifying the
peer representations to expose, their insertion points in local computation,
the aggregation operator, and the adaptation strategy for learning the
resulting interaction. We leave these choices open here and investigate them
through the controlled architecture study in
Section~\ref{sec:interaction_discovery}.

\subsection{Discovering the Interaction Architecture}
\label{sec:interaction_discovery}

\begin{figure}[!t]
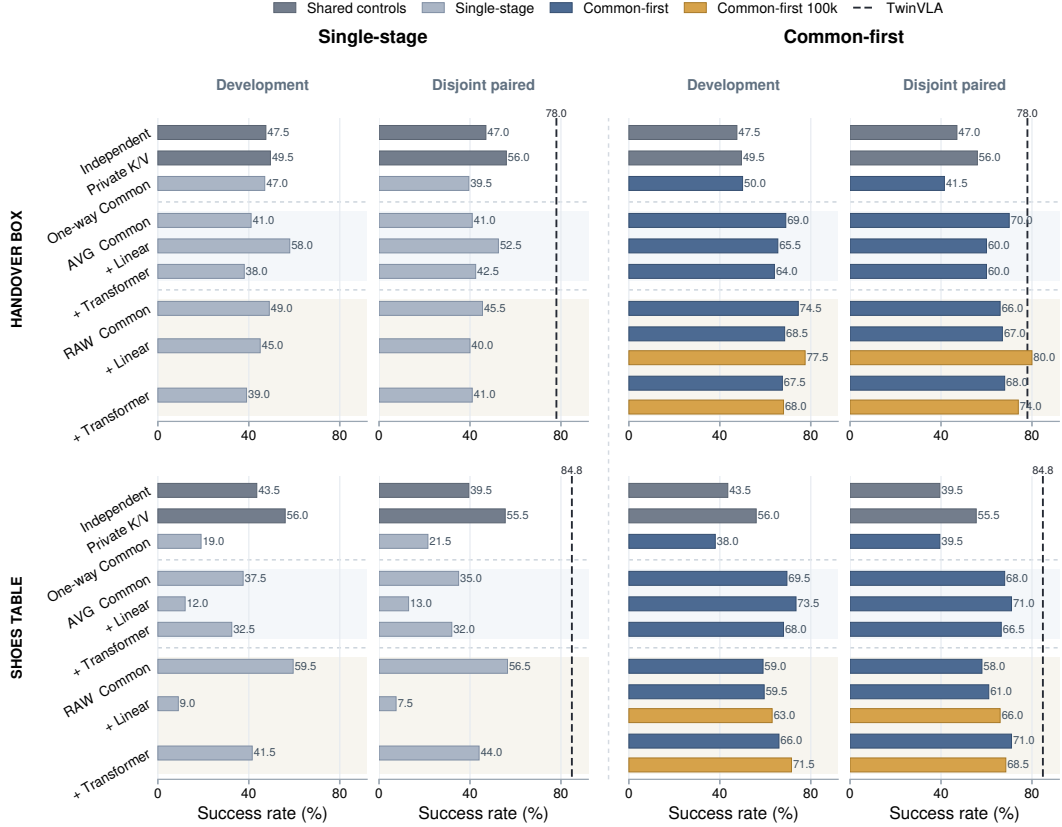

\centering
{\raggedright\small\bfseries (a) Search-space construction\par}
\input{figures/interaction_design_space}
\par\smallskip
{\raggedright\small\bfseries (b) Controlled study results\par}
\input{figures/interaction_validation200}
\caption{Discovering the VLA-Fabric interaction architecture.
(a) Search space over interaction entry, operator, and adaptation.
(b) Controlled results on Handover Box and Shoes Table.
Gold: 100k RAW follow-up; dashed: TwinVLA.}
\label{fig:derivation}
\label{fig:interaction_full_matrix}
\end{figure}

To study interaction between complete VLA agents, we expose the cross-arm
dependencies of a centralized policy while preserving its single-arm priors.
TwinVLA provides two identifiable pretrained SingleVLA branches
\citep{twinvla}, which we separate into local policies and reconnect through
explicit representation interfaces.

Figure~\ref{fig:derivation}(a) organizes the design space around three
decisions: where policies interact, how peer information is combined,
and how the resulting interaction is adapted.
\textbf{Interaction entry.} Candidates range from Independent policies and
Private K/V exchange to additional Common and action-side interaction.
\textbf{Interaction operator.} Common exchange uses one-way transfer,
symmetric averaging (AVG), or operator-level synchronization (RAW);
action-side interaction uses Linear or Transformer fusion.
\textbf{Interaction adaptation.} Single-stage enables the candidate
architecture from the start of fine-tuning; Common-first initializes it
from a Common-only policy before introducing the remaining interactions.

Nine Single-stage designs and seven matched Common-first variants per task
yield 32 runs on Handover Box and Shoes Table.
Checkpoints are selected on 200 development conditions and evaluated on
200 disjoint paired conditions, as summarized in
Figure~\ref{fig:interaction_full_matrix}(b).

\noindent\textbf{Where to interact.}
Private K/V alone improves over Independent policies on both tasks, and
richer configurations obtain further gains from Common and action-side
interaction. Although additional inputs do not always help, these comparisons
support complementary interaction across policy stages rather than one
insertion point for every coordination requirement.

\noindent\textbf{How to interact.}
The tasks impose different coordination demands, against which the operator
choices expose different advantages. Symmetric Common exchange improves over
one-way transfer in the Common-first comparisons, while action-side fusion
provides further gains in several configurations. Neither AVG versus RAW
nor Linear versus Transformer yields a consistent winner across both tasks.
These observations motivate retaining complementary interaction functions
that policies can exploit as coordination demands change, rather than
defining VLA-Fabric around one task's peak configuration.

\noindent\textbf{How to adapt.}
Common-first consistently improves the matched configurations, showing that
interaction structure and its adaptation should be considered jointly when
learning richer cross-policy coordination.
The interaction functions and their adaptation recipe thus play distinct
roles. The former specify which peer representations enter local computation,
how they are combined, and where they condition action generation.
The latter determines how policies learn to exploit this information during
task adaptation.

\noindent\textbf{Selecting a reusable realization.}
The targeted longer-run comparison finds no consistent advantage for
Transformer action fusion over the simpler Linear alternative.
We therefore carry forward RAW Common, Private K/V, and Linear action
interaction as the canonical SingleVLA-Fabric realization.
This choice yields a compact realization that retains the three complementary
coordination functions used in the subsequent transfer and scaling studies.

\subsection{From Architecture Search to Functional Interaction}
The architecture study motivates three complementary interaction roles.
\textbf{Shared semantic context} combines Common representations into
team-level context.
\textbf{Peer latent context} exposes peer K/V to local queries, allowing each
receiver to select information relevant to its own computation.
\textbf{Action intent} conditions local action generation on the peer's
action-side state.
In the selected SingleVLA realization, these roles correspond to RAW Common,
Private K/V, and Linear fusion of final \mbox{readouts}.
They define coordination functions rather than fixed tensor locations;
their formal operators are given in Appendix~\ref{app:functional_interface}.
Together, these three roles establish the core VLA-Fabric interaction
paradigm in the SingleVLA realization.

\section{VLA-Fabric Across VLA Architectures}
\label{sec:cross_backbone}

\subsection[Re-instantiating VLA-Fabric in pi0.5]{Re-instantiating VLA-Fabric in $\pi_{0.5}$}

\begin{figure}[!htb]
\centering
\input{figures/cross_backbone_transfer}
\end{figure}

The next question is whether the interaction functions discovered in
SingleVLA remain useful beyond its particular computational structure.
We use $\pi_{0.5}$ \citep{pi05} because its multimodal processing and iterative
Action Expert differ from SingleVLA's VLM-readout-conditioned action head.
We re-instantiate VLA-Fabric in this architecture, preserving complete local
policies and mapping the three coordination roles to architecture-appropriate
computation.

Figure~\ref{fig:cross_backbone} aligns the two realizations.
SingleVLA-Fabric combines Common and Private K/V within the Eagle2-based VLM,
then fuses final readouts before local action generation.
The fused readout conditions each arm's separate local action sampler.
In $\pi_{0.5}$, shared context is formed through Common multimodal prefix
fusion, while Private Prefix K/V supplies peer information to local
Action Expert queries.
Action interaction moves inside the Action Expert, where gated residual
peer attention conditions the evolving action states.
The same three coordination roles are therefore preserved through
architecture-specific realizations; detailed implementations are given in
Appendix~\ref{app:transfer_implementation}.

\subsection{Cross-Architecture Results and Interaction Use}
Table~\ref{tab:cross_backbone}A presents a four-task bimanual capability
comparison spanning Handover Box, Shoes Table, Handover Mic, and Scan Object.
Despite the architectural differences between Eagle2-based SingleVLA and
$\pi_{0.5}$, both realizations support cooperation through the same
interaction organization.
The $\pi_{0.5}$ results provide a second effective realization of the
architecture discovered in Section~\ref{sec:interaction}.
Table~\ref{tab:cross_backbone}B examines interaction use within each trained
policy.

\begingroup
\setlength{\intextsep}{5pt}
\begin{table}[H]
\centering\small
\caption{Bimanual capability and fixed-weight interaction removals (success, \%).}
\label{tab:cross_backbone}
\label{tab:functional_diagnostics}
\setlength{\tabcolsep}{9pt}
\renewcommand{\arraystretch}{0.95}
\begin{tabular}{llrrrr}
\specialrule{\heavyrulewidth}{0pt}{0pt}
\rowcolor{fabricpale}
\multicolumn{6}{l}{\color{black}\textbf{A. Bimanual capability}}\\
Policy & Setting & Box & Shoes & Mic & Scan\\
\midrule
RDT-1B \citep{rdt1b} & ICLR 2025 & 69.4 & 66.0 & 90.0 & 4.0\\
$\pi_0$ \citep{pi0} & RSS 2025 & 78.8 & 82.4 & \textbf{98.0} & 18.0\\
TwinVLA \citep{twinvla} & ICLR 2026 & 78.0 & 84.8 & 84.0 & 11.0\\
\cmidrule{1-6}
SingleVLA-Fabric & Full & 80.0 & 70.0 & 84.0 & 5.5\\
$\pi_{0.5}$-Fabric & Full & \textbf{80.5} & \textbf{94.5} & 97.5 & 33.0\\
\cmidrule{1-6}
Shared $\pi_{0.5}$-Fabric & Full & 69.0 & 78.0 & 97.0 & \textbf{40.0}\\
\arrayrulecolor{fabricrule}\specialrule{0.7pt}{3pt}{0pt}\arrayrulecolor{black}
\rowcolor{fabricmint}
\multicolumn{6}{l}{\color{black}\textbf{B. Fixed-weight interaction removals}}\\
Policy & Inference setting & Box & Shoes & Mic & Scan\\
\midrule
% Expand the row fragment before TeX starts the next alignment cell.
\csname @@input\endcsname tables/table_functional_diagnostics.tex
\bottomrule
\end{tabular}
\end{table}

\endgroup

\textbf{Full} enables all three interaction functions, while \textbf{All-off}
removes them at inference from a trained Full policy.
\textbf{Independent} is instead trained and evaluated without interaction.
Fixed-weight removals show substantial use of peer information, with the
relative importance of each function varying across coordination demands.
Action-side interaction is especially important for $\pi_{0.5}$ on Box,
whereas peer latent context plays a larger role on Mic and Scan.
Shoes remains comparatively robust to individual removals, while SingleVLA
is strongly affected across its evaluated tasks.
No single interaction function dominates across all settings, leaving room
to optimize when, where, and how strongly peer information is used.
The Shared $\pi_{0.5}$-Fabric setting jointly fine-tunes the local policies
and their interaction modules across all four tasks, using one checkpoint
per role for cross-task reuse.

\FloatBarrier
\section{Scaling VLA-Fabric Beyond Bimanual Manipulation}
\label{sec:simulation}

\begingroup
\setlength{\intextsep}{5pt}
\begin{figure}[!htbp]
\centering
\setlength{\parskip}{0pt}
\IfFileExists{figures/multiarm_extensibility.pdf}{%
  \includegraphics[width=0.92\linewidth]{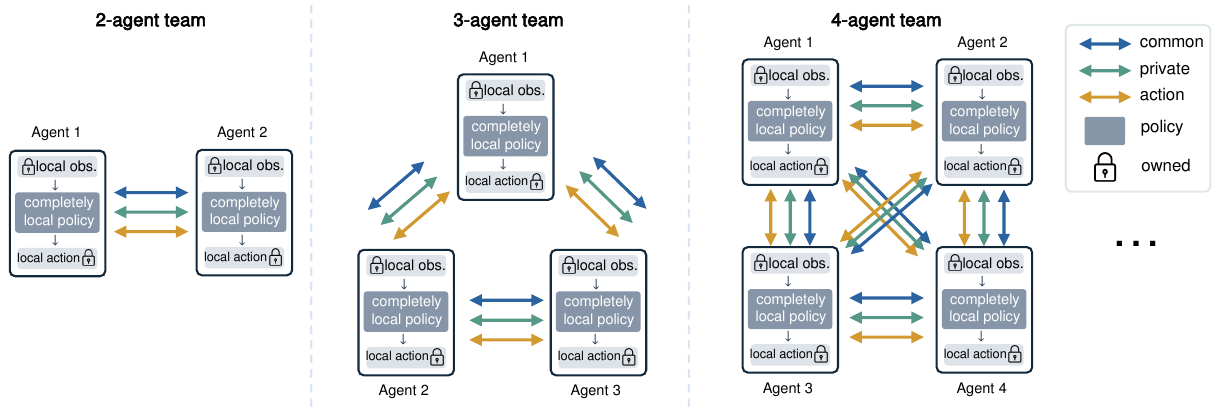}%
}{%
  \includegraphics[width=0.92\linewidth]{figures/multiarm_extensibility.png}%
}
\par
\begin{minipage}[t]{0.237\linewidth}
  \centering\small\textbf{3-arm} Stack Cube\par\smallskip
  \includegraphics[width=\linewidth]{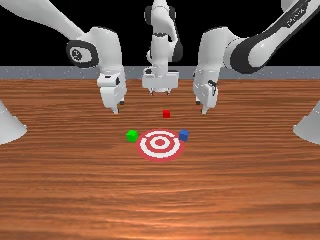}
\end{minipage}\hfill
\begin{minipage}[t]{0.237\linewidth}
  \centering\small\textbf{3-arm} Frame-over-Peg\par\smallskip
  \includegraphics[width=\linewidth]{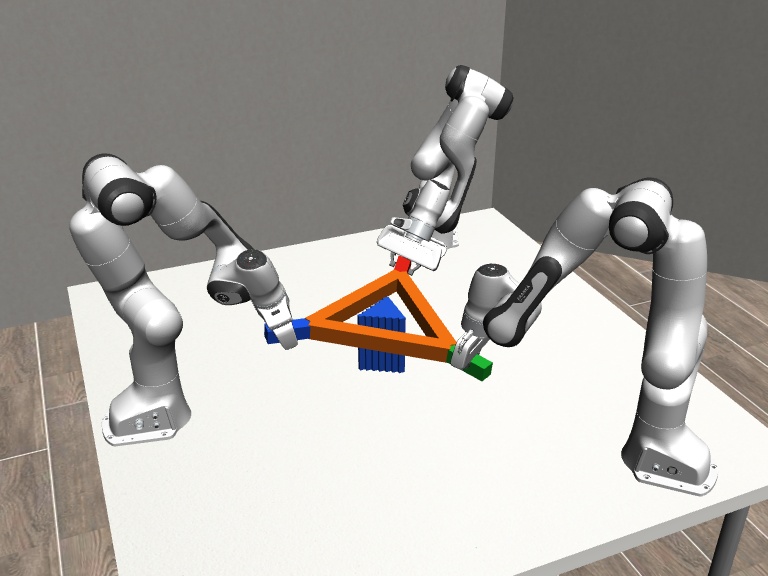}
\end{minipage}\hfill
\begin{minipage}[t]{0.237\linewidth}
  \centering\small\textbf{4-arm} Frame-over-Peg\par\smallskip
  \includegraphics[width=\linewidth]{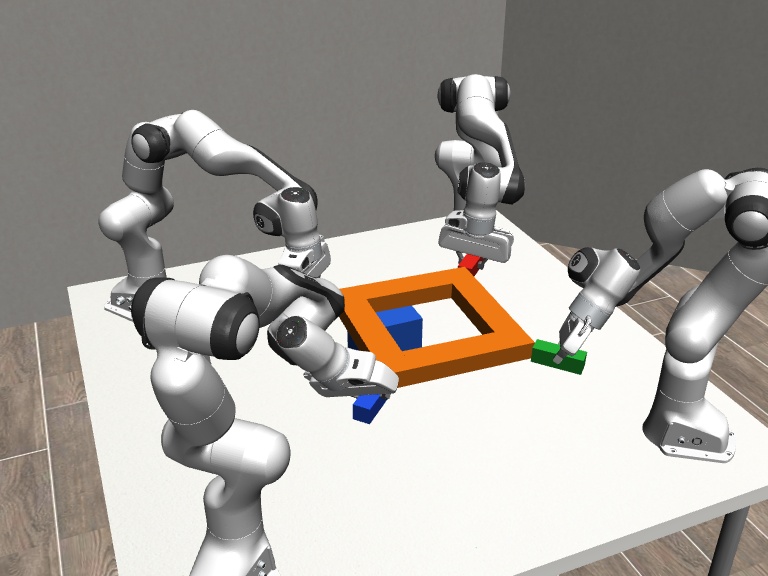}
\end{minipage}\hfill
\begin{minipage}[t]{0.237\linewidth}
  \centering\small\textbf{4-arm} Arch Building\par\smallskip
  \includegraphics[width=\linewidth]{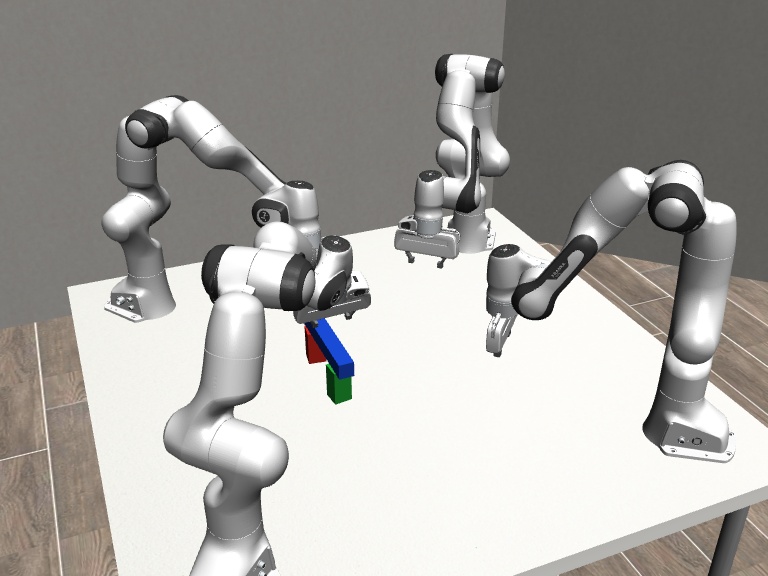}
\end{minipage}
\caption{Two- to four-agent VLA-Fabric teams and larger-team simulation
tasks (Table~\ref{tab:larger_team_main}).}
\label{fig:task_overview}
\end{figure}

\endgroup

VLA-Fabric aims to move beyond fixed bimanual policies toward multi-arm
systems composed of complete VLA agents.
Having established the interaction architecture and re-instantiated it across
VLA families, we now examine its ability to organize larger collaborative
teams.
We extend the same peer-interaction structure to three and four policies,
each retaining its own local action generation and receiving information
from an expanded peer set.
The local action heads retain their dimensions as the team gains additional
roles.
We evaluate three-arm Stack Cube (Stack3) from RoboFactory \citep{robofactory}
and introduce three- and four-arm Frame-over-Peg (Frame3/4) and four-arm
Arch Building (Arch4) for more demanding multi-arm coordination
(Figure~\ref{fig:task_overview}).

\begingroup
\setlength{\columnsep}{12pt}
\setlength{\intextsep}{5pt}
\begin{wraptable}{r}{0.37\textwidth}
\compactwrapcaption
\caption{Larger-team success (\%).}
\label{tab:larger_team_main}
\small
\setlength{\tabcolsep}{4pt}
\renewcommand{\arraystretch}{1.06}
\begin{tabular*}{\linewidth}{@{\extracolsep{\fill}}lrr@{}}
\toprule
Task & Independent & Full\\
\midrule
Stack3 & 90.5 & \textbf{91.0}\\
Frame3 & 67.5 & \textbf{79.5}\\
Frame4 & 9.5 & \textbf{89.5}\\
Arch4 & 0.0 & \textbf{5.0}\\
\bottomrule
\end{tabular*}
\end{wraptable}

Table~\ref{tab:larger_team_main} suggests that the benefit of explicit
interaction depends on how local actions constrain one another.
Stack3 largely decomposes into role-specific pick-and-place behaviors,
allowing the constituent VLAs to perform well with little measured benefit
from interaction.
In the frame tasks, several arms continuously support and align a shared
object, so each arm's motion changes the conditions under which its peers act.
The stronger Full results are consistent with explicit interaction helping
policies accommodate these mutual dependencies.
VLA-Fabric thus supports both relatively decomposable teams and tightly
coupled manipulation, with the value of peer information shaped by the task.
Arch4 highlights a complementary challenge: its long-horizon, high-precision
assembly combines random object displacement with opportunities for inter-arm
collisions.
Full achieves some successful completions while Independent achieves none
in the reported evaluations, but reliable assembly remains difficult.
Organizing policies through interaction can support coordination without
automatically resolving the constituent VLAs' planning and precision-control
limitations.
\par\endgroup

\FloatBarrier
\section{VLA-Fabric in Real-World Multi-Arm Systems}
\label{sec:real}

Having established the interaction architecture, transferred it across VLA
families, and extended it to larger simulated teams, we now evaluate
VLA-Fabric in physical multi-arm systems.
The study covers four three- and four-arm tasks involving parallel placement,
object relay, and shared-object manipulation.
We collect 50 demonstrations per task and evaluate Full and independently
trained local $\pi_{0.5}$ policies over 25 physical trials per policy setting.
\textbf{Basket3} has three arms place vegetables into a shared basket.
In \textbf{Relay3}, one arm positions the receiving target, a second hands
over an object, and the third receives and places it in the target.
\textbf{Relay4} uses two arm pairs, each handing over and depositing vegetables.
\textbf{Frame4} requires four arms to jointly lift, align, and lower a
frame over a peg.

\begin{figure}[!htbp]
\centering
\begingroup
\setlength{\unitlength}{0.16\linewidth}
\newcommand{\contactframe}[1]{%
  \begin{picture}(1,0.75)%
    \put(0,0){\includegraphics[width=\unitlength]{#1}}%
  \end{picture}%
}
\newcommand{\contactrow}[2]{%
  \hbox{%
    \begin{picture}(0.25,0.75)%
      \put(0.12,0.375){\makebox(0,0){\rotatebox{90}{\scriptsize\textbf{#1}}}}%
    \end{picture}%
    #2%
  }%
}
\IfFileExists{figures/real_robot_sequences.pdf}{%
  \includegraphics[width=\linewidth]{figures/real_robot_sequences.pdf}%
}{%
  \vbox{\offinterlineskip
    \contactrow{Basket3}{%
      \contactframe{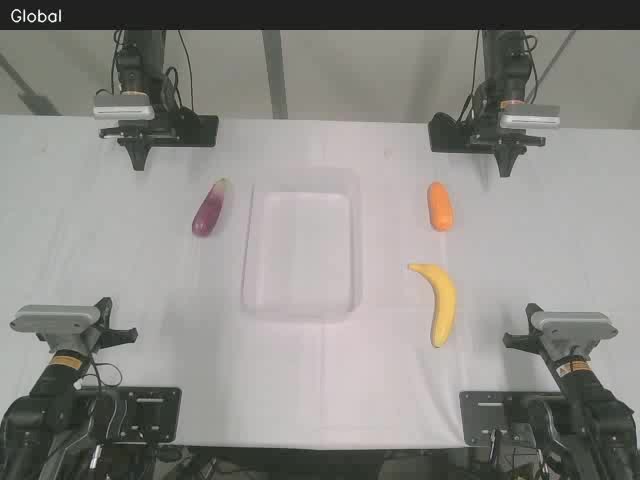}%
      \contactframe{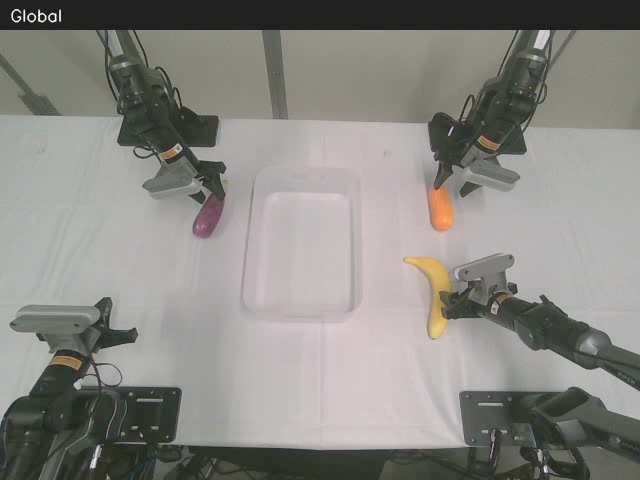}%
      \contactframe{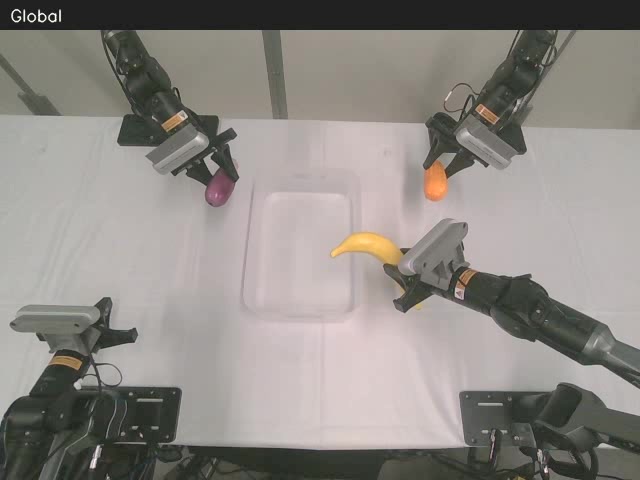}%
      \contactframe{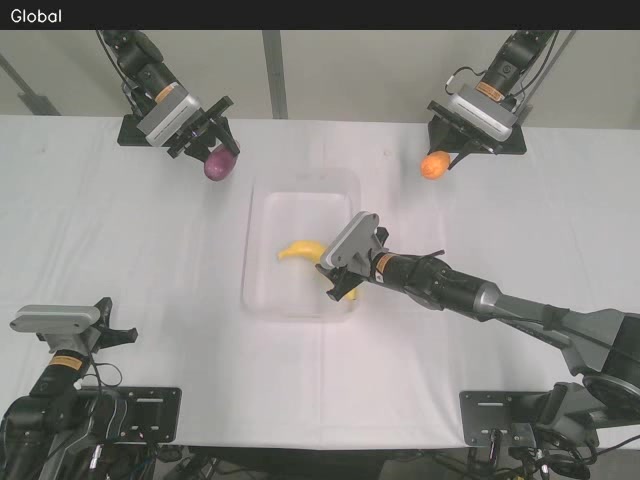}%
      \contactframe{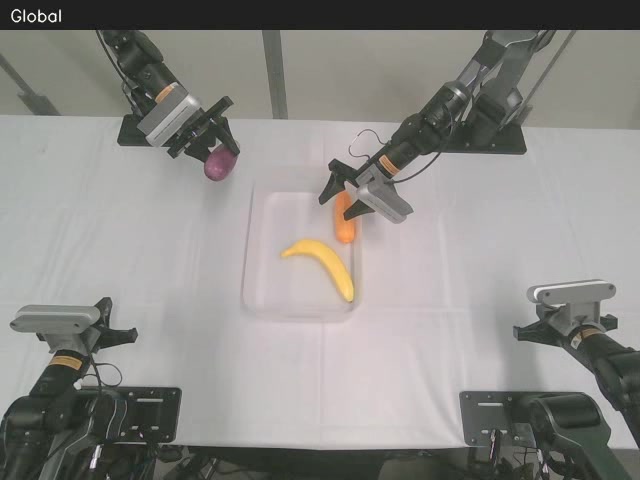}%
      \contactframe{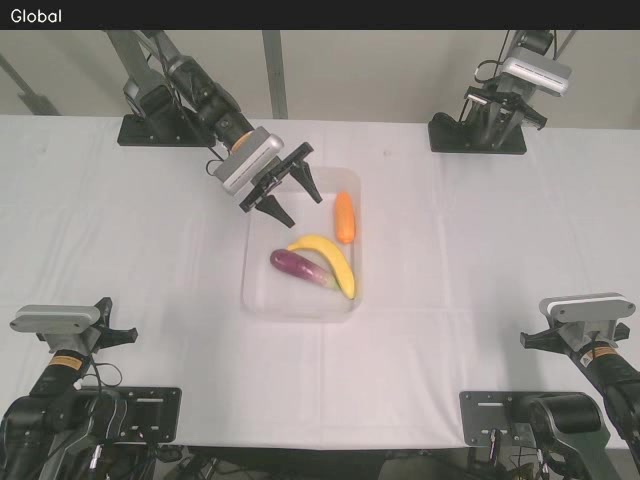}%
    }%
    \contactrow{Relay3}{%
      \contactframe{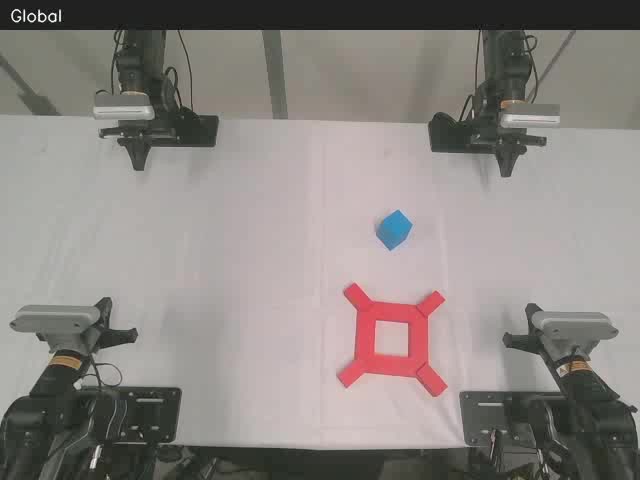}%
      \contactframe{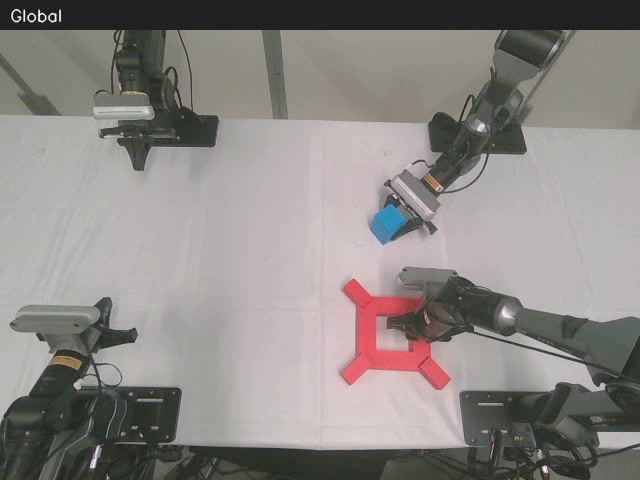}%
      \contactframe{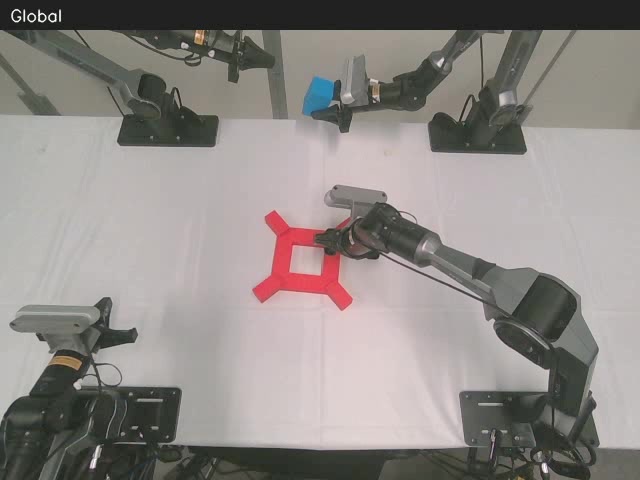}%
      \contactframe{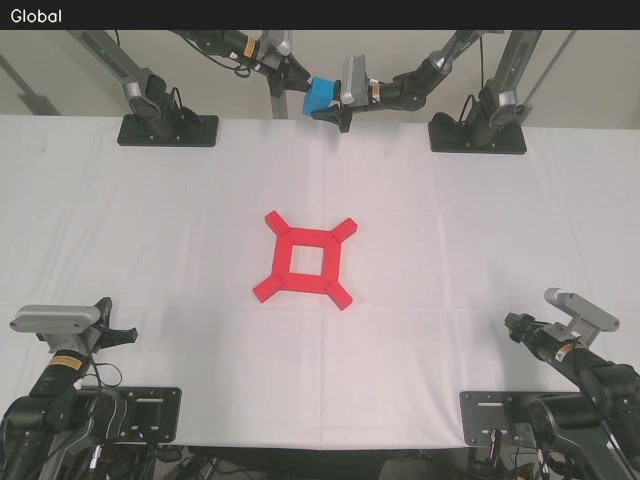}%
      \contactframe{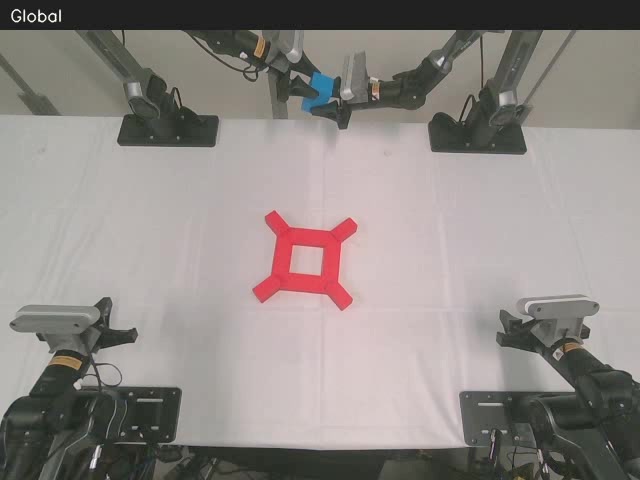}%
      \contactframe{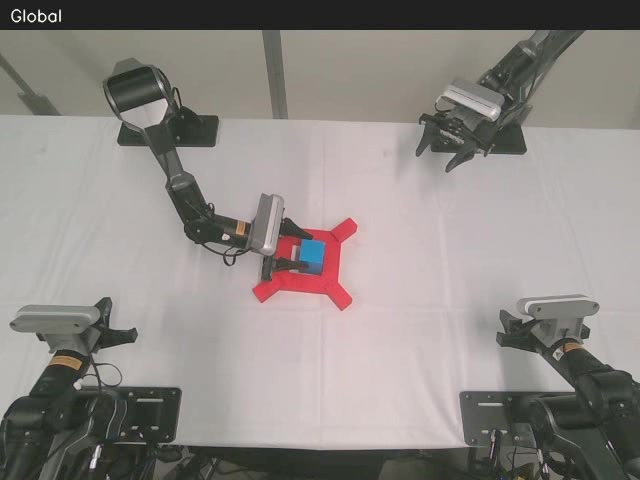}%
    }%
    \contactrow{Relay4}{%
      \contactframe{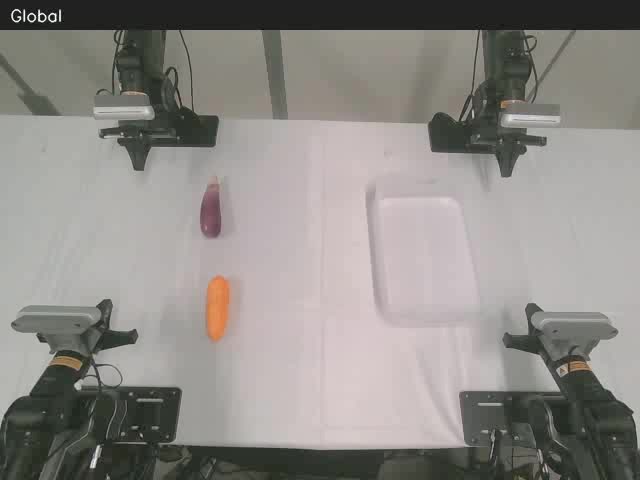}%
      \contactframe{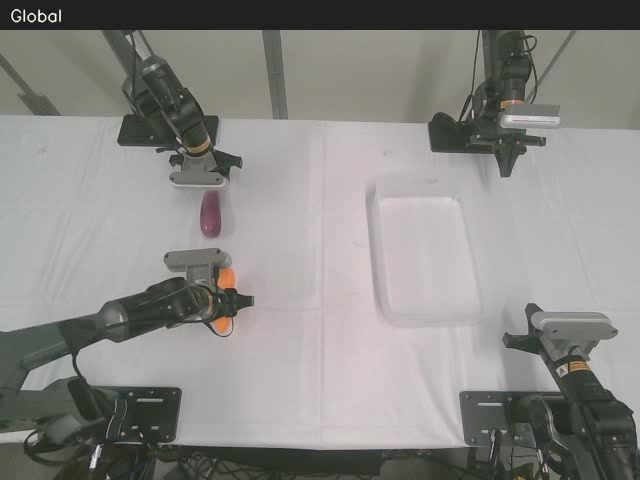}%
      \contactframe{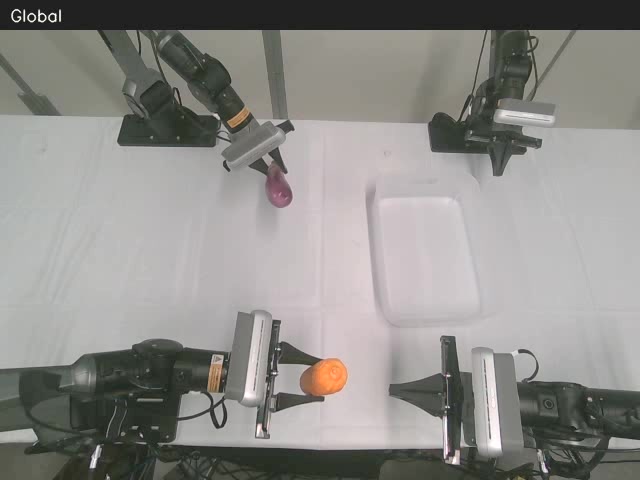}%
      \contactframe{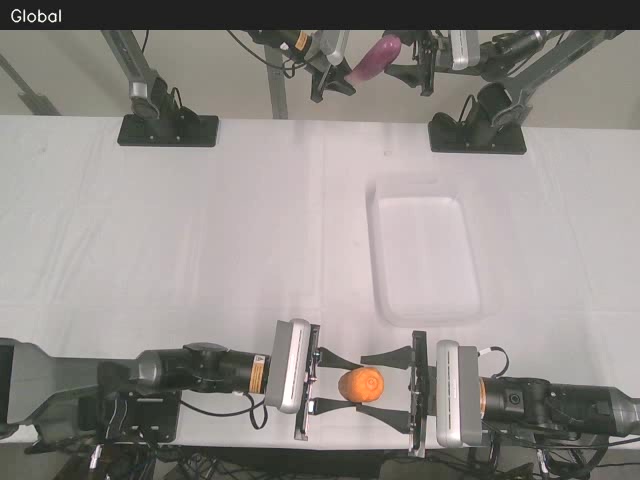}%
      \contactframe{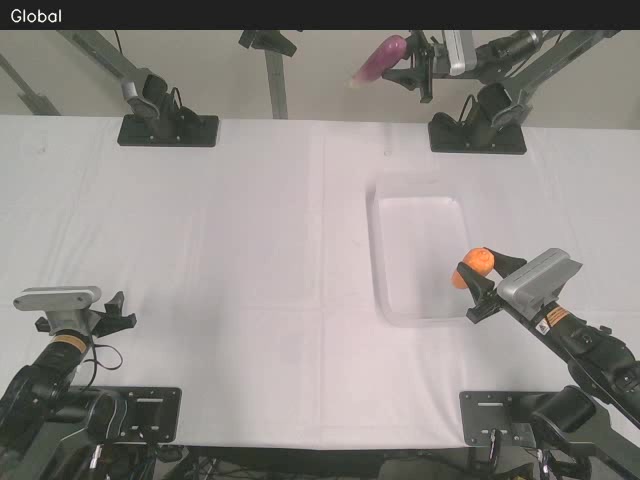}%
      \contactframe{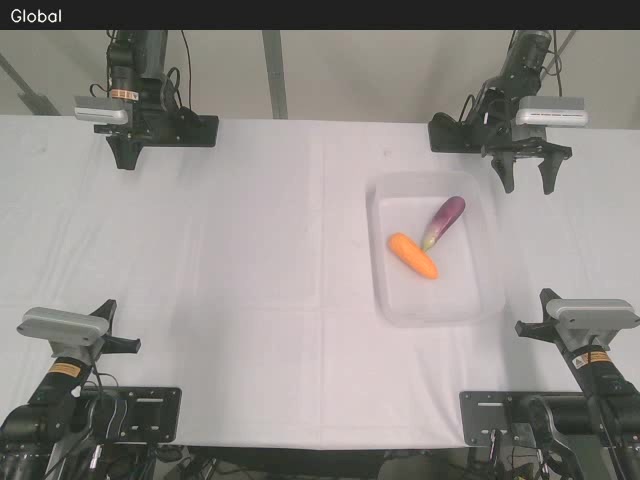}%
    }%
    \contactrow{Frame4}{%
      \contactframe{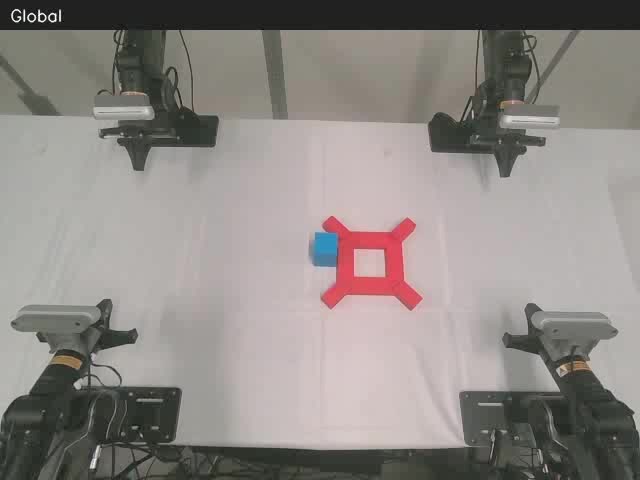}%
      \contactframe{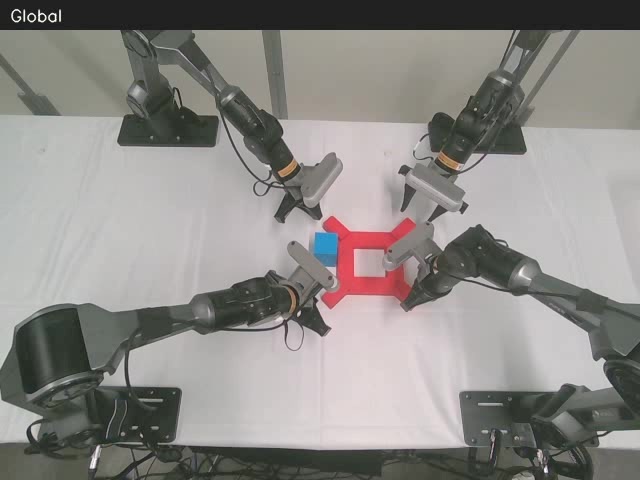}%
      \contactframe{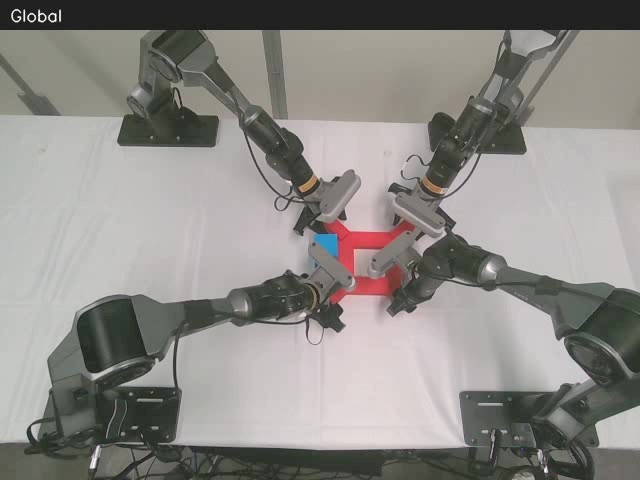}%
      \contactframe{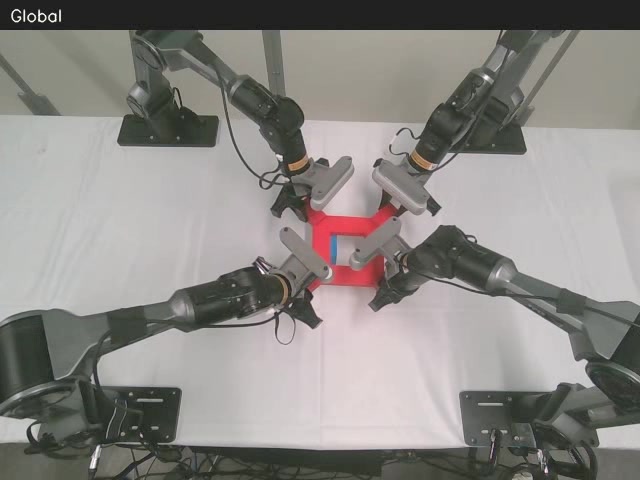}%
      \contactframe{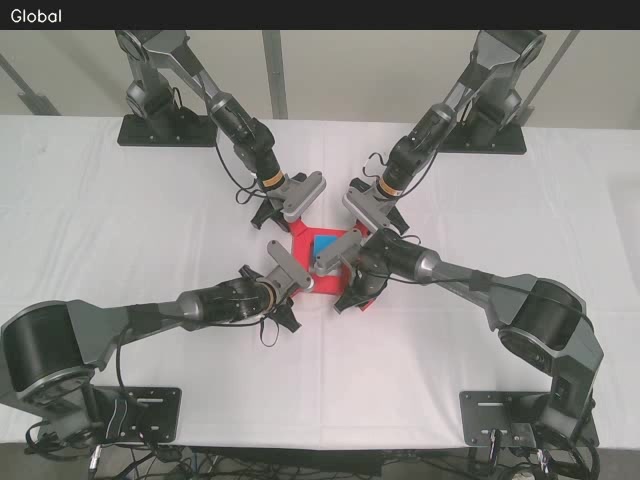}%
      \contactframe{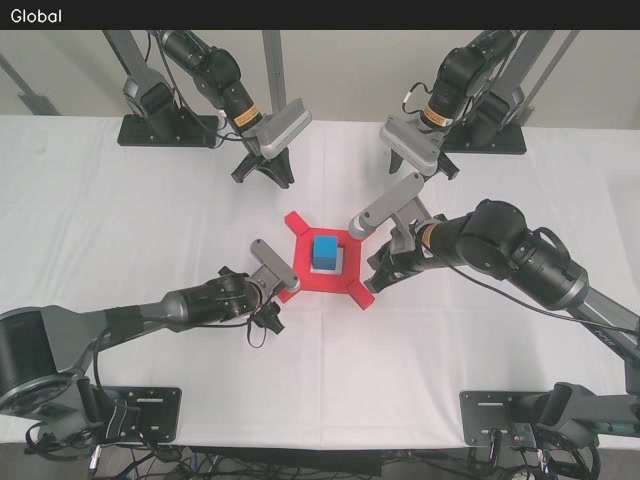}%
    }%
  }
}
\endgroup
\caption{Representative physical executions of the four real-world tasks;
each row shows six chronological global-camera frames.}
\label{fig:real_robot_sequences}
\end{figure}

\begingroup
\setlength{\columnsep}{12pt}
\setlength{\intextsep}{5pt}
\Needspace{10\baselineskip}
\begin{wraptable}{r}{0.37\textwidth}
\compactwrapcaption
\caption{Physical success (\%).}
\label{tab:physical_main}
\small
\setlength{\tabcolsep}{4pt}
\renewcommand{\arraystretch}{1.06}
\begin{tabular*}{\linewidth}{@{\extracolsep{\fill}}lrr@{}}
\toprule
Task & Independent & Full\\
\midrule
Basket3 & 72.0 & \textbf{88.0}\\
Relay3 & 16.0 & \textbf{56.0}\\
Relay4 & 40.0 & \textbf{72.0}\\
Frame4 & 36.0 & \textbf{60.0}\\
\bottomrule
\end{tabular*}
\end{wraptable}

VLA-Fabric Full improves over Independent on all four physical tasks by
16--40 percentage points (Table~\ref{tab:physical_main}).
The pattern is consistent with the simulation study: Basket3 is comparatively
decomposable and already supports strong local execution.
Relay3, Relay4, and Frame4 impose tighter dependencies through handover timing
or shared-object control and show larger benefits from explicit interaction.
In a relay, the handing arm must maintain support until the receiving arm
can take over. In Frame4, each arm's motion affects the frame supported by
all four arms. Such dependencies require coordination during action
execution, beyond assigning each arm a role or target. The physical study
therefore examines whether complete local policies can accommodate these
shared constraints under perception and control uncertainty.
The results demonstrate that the multi-VLA organization remains effective
under these conditions, while reliable execution of the more coupled behaviors
remains an important direction for improvement.
Across simulation and hardware, the recurring pattern suggests that
interaction benefits reflect physical coupling among agents, beyond arm
count alone.
\par\endgroup

\section{Discussion and Limitations}
\label{sec:limitations}

VLA-Fabric makes a complete local VLA an explicit organizational unit of
multi-arm coordination.
Local perception and action ownership are preserved, while collaboration is
learned through representation exchange across policy boundaries.
The same coordination roles can be re-instantiated across VLA architectures,
and additional policies can form larger teams without redefining a global
joint action space.
This separates progress in constituent models from progress in team
organization, allowing the two to advance together.
Stronger VLAs can improve the execution units, while interaction mechanisms
determine which information is shared, when it enters policy computation,
and with which peers.
Across the evaluated architectures, tasks, and team sizes, this common
organization supports reuse of the coordination principle while allowing
the underlying policy computation to change.

Our experiments use homogeneous teams with fixed roles and up to
four arms, leaving changing team compositions and heterogeneous VLAs as
directions for broader composability.
Broader composition will require interfaces that accommodate differences in
representation and action generation while maintaining clear local policy
boundaries.
Such interfaces must align the meaning and timing of exchanged information
across policies with different internal computations.
The current interaction structure is static; adapting content, topology, and
frequency to task phase and agent participation could make coordination more
selective.
Long-horizon, high-precision manipulation remains difficult, as illustrated
by Arch Building, motivating complementary improvements in planning and
control of the constituent VLAs.
Larger teams also motivate more selective representation exchange and
interaction mechanisms that accommodate changing role assignments and
available communication resources.

\section{Conclusion}
\label{sec:conclusion}

In this paper, we introduced VLA-Fabric to address how multiple complete VLA
policies can collaborate as a coordinated system for increasingly demanding
multi-arm manipulation.
Each arm-local VLA remains a complete execution unit, preserving its own
perception and action generation while acquiring collaborative capability
through learned representation interaction.
We established this organization through systematic exploration of interaction
entry, aggregation, and adaptation, then re-instantiated its coordination
functions in a structurally different VLA.
Extending the same principle to three- and four-agent teams in simulation and
physical systems completes a framework for discovering, transferring, and
scaling inter-VLA coordination.
This organization allows individual VLA capability and multi-VLA coordination
to advance together, connecting improvements in local models with the
mechanisms through which they collaborate.
Beyond the operators studied here, it opens directions in coordination-aware
model design, adaptive representation exchange, communication-efficient
deployment, and heterogeneous VLA teams.
VLA-Fabric thereby offers a concrete step from increasingly capable individual
embodied models toward networked embodied intelligence.

\section*{AI Use Statement}

Generative AI tools were used to assist research planning, literature organization, manuscript structuring, language editing, and code-oriented document preparation. The authors verified all cited sources, implementations, experimental protocols, numerical results, analyses, and scientific claims, and take full responsibility for the submission.

\section*{Reproducibility Statement}

Appendices~\ref{app:protocols}--\ref{app:multiarm_recipe} mirror
the discovery, transfer, and larger-team studies, documenting implementation,
selection, complete results, and evaluation boundaries. The
\href{https://nebulis-lab.com/VLA-Fabric/}{project website}
links the \href{https://github.com/NEBULIS-Lab/VLA-Fabric}{code release},
including both VLA realizations and 2/3/4-agent workflows. Pretrained weights
and datasets are obtained separately as documented in the release.

\setlength{\bibsep}{4pt plus 1pt minus 1pt}
\bibliography{refs}
\bibliographystyle{iclr2027_conference}

\clearpage
\appendix
\section*{Appendix}
\setcounter{table}{0}
\counterwithin*{table}{section}
\renewcommand{\thetable}{\thesection\arabic{table}}
\renewcommand{\theHtable}{appendix.\thesection.\arabic{table}}
\setcounter{figure}{0}
\counterwithin*{figure}{section}
\renewcommand{\thefigure}{\thesection\arabic{figure}}
\renewcommand{\theHfigure}{appendix.\thesection.\arabic{figure}}
\numberwithin{equation}{section}
% Supplementary sections mirror the three main evidence stages.
\raggedbottom
\FloatBarrier
Appendices~\ref{app:protocols}--\ref{app:multiarm_recipe} provide supporting
formulations, experimental settings, and additional results for interaction
discovery, cross-architecture transfer, and larger-team scaling, respectively.
\section{Interaction Architecture Discovery}
\label{app:protocols}

\subsection{SingleVLA Decomposition and Design Space}
\label{app:discovery_design}

TwinVLA's two pretrained SingleVLA branches expose identifiable cross-arm
dependencies while preserving each arm's pretrained prior \citep{twinvla}.
We separate these branches into complete local policies and vary the
representations through which they interact. All candidates originate from
the same pretrained SingleVLA policies, using an Eagle2 VLM-based backbone.
Table~\ref{tab:app_design_space} defines the nine interaction designs.

One-way Common shares the Common keys and values from one designated arm.
AVG Common averages the two agents' Common keys and values before they
condition private attention. RAW Common synchronizes Common computation at operator boundaries,
before the agents proceed with their local computation.
Both symmetric variants retain Private K/V interaction.
The Linear and Transformer variants additionally combine role-ordered
action-side readouts before each arm's local action head.
Thus aggregation, receiver-specific latent conditioning, and late-action
capacity can be varied without changing local observation or action ownership.

\begin{table}[!htbp]
\centering\small
\caption{Interaction design space. Each design preserves complete local VLA policies.}
\label{tab:app_design_space}
\begin{tabular*}{\linewidth}{@{\extracolsep{\fill}}llll@{}}
\toprule
Design & Common interaction & Private K/V & Action interaction\\
\midrule
I0 & None & Off & None\\
I1 & None & On & None\\
I2 & One-way & On & None\\
I3-AVG & Averaged K/V & On & None\\
I3-RAW & Operator-level & On & None\\
I4-AVG & Averaged K/V & On & Linear\\
I4-RAW & Operator-level & On & Linear\\
I5-AVG & Averaged K/V & On & Transformer\\
I5-RAW & Operator-level & On & Transformer\\
\bottomrule
\end{tabular*}
\end{table}

\subsection{Functional Formulation of VLA-Fabric}
\label{app:functional_interface}

Let $\mathcal N(i)$ denote agent $i$'s incoming peers and $k$ a layer or
interaction stage; the planning-round index is omitted below.
The three functions condition local action generation at different
computational boundaries.

\paragraph{Shared semantic context.}
Agent $j$ exposes its Common-owned representation $h_{j,\mathrm{sem}}^k$.
The symmetric operator $\operatorname{Agg}_{\mathrm{sem}}^k$ combines these
representations into shared context while private observations remain local:
\begin{equation}
\widetilde h_{i,\mathrm{sem}}^k
 = \operatorname{Agg}_{\mathrm{sem}}^k
 \bigl(\{h_{j,\mathrm{sem}}^k:j\in\mathcal N(i)\cup\{i\}\}\bigr).
\label{eq:semantic}
\end{equation}
This expression describes the symmetric Common realizations; the one-way
candidate instead restricts the direction of context transfer.

\paragraph{Peer latent context.}
Agent $i$ retains its local query $Q_i^k$ and local keys and values
$K_i^k,V_i^k$. The role-ordered peer keys and values
$\widehat K_{\mathcal N(i)}^k,\widehat V_{\mathcal N(i)}^k$ augment this context.
Attention produces the receiver-specific latent context $c_i^k$:
\begin{equation}
c_i^k = \operatorname{Attn}\bigl(Q_i^k,
 [K_i^k,\widehat K_{\mathcal N(i)}^k],
 [V_i^k,\widehat V_{\mathcal N(i)}^k]\bigr).
\label{eq:latent}
\end{equation}
Brackets denote concatenation along the token dimension.
Each receiver selects relevant peer information through its own local query.

\paragraph{Action intent.}
Each $z_j$ represents agent $j$'s action-side state, a final readout in the
selected SingleVLA implementation. Concatenation follows a fixed role order,
and a receiver-specific fusion operator $\phi_i^{\mathrm{act}}$ produces
$\widetilde z_i$ for local action generation:
\begin{equation}
\widetilde z_i = \phi_i^{\mathrm{act}}\bigl(
 \operatorname{Concat}^{\mathrm{role}}_{j\in\mathcal N(i)\cup\{i\}}z_j\bigr).
\label{eq:action_intent}
\end{equation}
The selected realization uses Linear fusion followed by the local
flow-matching action head. Equations~\ref{eq:semantic}--\ref{eq:action_intent}
describe the three functional roles through their respective interaction operators.
Section~\ref{sec:cross_backbone} re-instantiates them within a different VLA
computation.

\subsection{Experimental Setup}
\label{app:controlled_study}

The controlled study uses ALOHA Handover Box and Shoes Table.
Each task includes nine Single-stage designs and seven Common-first variants,
giving 32 model--task runs. Independent and Private K/V lack Common
interaction and serve as shared controls across the two adaptation settings.

Single-stage enables the target interaction architecture directly during
fine-tuning. Common-first initializes the policies with Common interaction
before introducing the remaining interaction functions.
The total adaptation budgets are 40k updates for Single-stage and 50k for
Common-first, including initialization. With these unequal budgets, the
comparison measures the sensitivity of each design to the complete adaptation recipe.

All variants freeze the vision encoder and adapt the language backbone,
action heads, and enabled interaction modules. Interaction operators are
evaluated in one differentiable training graph, with the native action
objective applied to every arm. A normalized description is
\begin{equation}
\mathcal L_{\mathrm{act}}=
\frac{1}{N}\sum_{i=1}^{N}
\mathcal L_{\mathrm{VLA}}\!\left(
\widehat a_i^{t:t+H-1},a_i^{t:t+H-1}\right).
\label{eq:action_loss}
\end{equation}
Here $H$ denotes the predicted action horizon. The implementation sums the
arm-local losses, differing by a constant scale at fixed team size.
The prediction target is unchanged across interaction designs.

Models are selected using 200 development conditions and evaluated on
200 disjoint paired conditions per task. The paired conditions are shared
across designs, allowing comparison under the same task initializations.
Success is the proportion of episodes satisfying the task's completion
criterion; episodes reaching the horizon without completion remain failures.

\subsection{Complete Architecture-Search Results}
\label{app:controlled_findings}

Table~\ref{tab:app_architecture_results} gives the complete numerical matrix
underlying the paired-condition panels in Figure~\ref{fig:interaction_full_matrix}.
Each row reports one design and adaptation choice on both tasks.
Common-first improves all seven matched designs, but the ranking of Common
aggregation and action operators varies across tasks.
The matrix therefore distinguishes access to peer information from the
policy's ability to learn useful conditioning on that information.

\begingroup
\small
\setlength{\LTleft}{0pt}
\setlength{\LTright}{0pt}
\setlength{\LTcapwidth}{\linewidth}
\begin{longtable}{@{}p{0.32\linewidth}p{0.26\linewidth}>{\raggedleft\arraybackslash}p{0.18\linewidth}>{\raggedleft\arraybackslash}p{\dimexpr0.24\linewidth-6\tabcolsep\relax}@{}}
\caption{Complete architecture-search results. Success rates (\%) use
200 paired conditions per task, disjoint from model selection.}
\label{tab:app_architecture_results}\\
\toprule
Design & Adaptation & Handover Box & Shoes Table\\
\midrule
\endfirsthead
\multicolumn{4}{l}{Table~\thetable\ (continued)}\\
\toprule
Design & Adaptation & Handover Box & Shoes Table\\
\midrule
\endhead
\midrule
\endfoot
\bottomrule
\endlastfoot
I0: Independent & Single-stage & 47.0 & 39.5\\
I1: Private K/V & Single-stage & 56.0 & 55.5\\
I2: One-way Common & Single-stage & 39.5 & 21.5\\
I3-AVG & Single-stage & 41.0 & 35.0\\
I3-RAW & Single-stage & 45.5 & 56.5\\
I4-AVG: Linear & Single-stage & 52.5 & 13.0\\
I4-RAW: Linear & Single-stage & 40.0 & 7.5\\
I5-AVG: Transformer & Single-stage & 42.5 & 32.0\\
I5-RAW: Transformer & Single-stage & 41.0 & 44.0\\
\midrule
I2: One-way Common & Common-first & 41.5 & 39.5\\
I3-AVG & Common-first & 70.0 & 68.0\\
I3-RAW & Common-first & 66.0 & 58.0\\
I4-AVG: Linear & Common-first & 60.0 & 71.0\\
I4-RAW: Linear & Common-first & 67.0 & 61.0\\
I5-AVG: Transformer & Common-first & 60.0 & 66.5\\
I5-RAW: Transformer & Common-first & 68.0 & 71.0\\
\end{longtable}
\endgroup

\subsection{Architecture Selection and SingleVLA Controls}
\label{app:longrun_selection}
\label{app:eagle_functional_controls}

\begin{wraptable}{r}{0.50\textwidth}
\compactwrapcaption
\caption{Extended-budget Common-first comparison, with 110k total updates
including initialization. Success rates (\%) use 200 paired conditions per task.}
\label{tab:app_extended_results}
\small
\setlength{\tabcolsep}{4pt}
\renewcommand{\arraystretch}{1.06}
\begin{tabular*}{\linewidth}{@{\extracolsep{\fill}}lrr@{}}
\toprule
Design & \shortstack{Handover\\Box} & \shortstack{Shoes\\Table}\\
\midrule
I4-RAW: Linear & 80.0 & 66.0\\
I5-RAW: Transformer & 74.0 & 68.5\\
\bottomrule
\end{tabular*}
\end{wraptable}

The extended comparison considers two preselected architecture families,
RAW Common with Linear or Transformer action interaction.
Table~\ref{tab:app_extended_results} compares their terminal models at the
same total budget. Added Transformer capacity does not provide a consistent
advantage across tasks, supporting Linear fusion as a compact common
realization for subsequent studies. This targeted follow-up reuses the
architecture-analysis conditions.

Independent in Table~\ref{tab:app_architecture_results} is trained without
all three interaction functions. In contrast, the SingleVLA removals in
Table~\ref{tab:cross_backbone} retain the trained Full weights and suppress
selected information only at inference. All-off disables Common aggregation,
Private K/V, and action fusion together. These removal results come from
a separate fixed-weight removal batch.
They test how trained policies use interaction; Independent tests what
policies learn without an interaction interface.

\FloatBarrier
\section{Cross-Architecture Transfer of VLA-Fabric}
\label{app:transfer}

\subsection[From SingleVLA to pi0.5]{From SingleVLA to $\pi_{0.5}$}
\label{app:transfer_implementation}

\paragraph{Action-generation boundary.}
Both VLA families generate continuous actions with flow matching
\citep{flowmatching}. SingleVLA uses an Eagle2-based VLM and a separate
DiT action head \citep{eagle2,twinvla}. Its backbone readout is fused once
before the two local action heads run their iterative samplers.
In $\pi_{0.5}$, peer-action interaction instead operates on the evolving
Action Expert suffix within each flow step.

\paragraph{Common and Private Prefix interaction.}
Each arm retains a complete $\pi_{0.5}$ policy with its own parameters,
proprioception, action-generation state, and output head.
Common interaction synchronizes the Common-owned multimodal PaliGemma
prefix at seven operator boundaries in each of 18 layers.
Private Prefix K/V remain associated with their sending role and augment
the context available to the receiver's Action Expert.
The receiver keeps its query local, so peer context is selected according
to its own evolving action computation. Raw private images are not exchanged.

\paragraph{Receiver-local action interaction.}
A separate peer-action attention branch adds a gated residual before the
Action Expert feed-forward block. For receiver $i$ and peer $j$, let $k$
index the expert layer and $s$ the flow step. Peer action-side keys and
values condition the local action-side query:
\begin{equation}
 r_i^{k,s}=\alpha_i^k W_i^k
 \operatorname{Attn}\!\left(Q_{i,\mathrm{act}}^{k,s},
 K_{j,\mathrm{act}}^{k,s},V_{j,\mathrm{act}}^{k,s}\right),
 \qquad h_i^{k,s}=h_{i,\mathrm{core}}^{k,s}+r_i^{k,s}.
\label{eq:pi_action_interaction}
\end{equation}
Here $h_{i,\mathrm{core}}^{k,s}$ is the receiver's core attention state;
$W_i^k$ and $\alpha_i^k$ are its output projection and scalar gate.
The gate is zero-initialized and learned without a hard bound, initially
preserving the core computation while allowing the interaction to develop.
Peer attention has a separate softmax from core attention and is refreshed
at every Action Expert layer and flow step.
The update conditions the receiver's evolving action representations throughout
the iterative generation process.

The correspondence with Appendix~\ref{app:functional_interface} is functional:
Common forms shared context, Private Prefix K/V provide receiver-specific
context, and the residual branch supplies action-stage interaction.
Each homogeneous team uses interaction parameters fine-tuned within its own
VLA architecture.

\subsection{Cross-Architecture Experimental Setup}
\label{app:cross_backbone_protocol}
\label{app:mic_matched}
\label{app:scan_object}

\begin{wraptable}{r}{0.51\textwidth}
\compactwrapcaption
\caption{Task-specific $\pi_{0.5}$ Full settings. Update counts include
all adaptation, including initialization where used.}
\label{tab:app_transfer_setup}
\small
\setlength{\tabcolsep}{2.5pt}
\renewcommand{\arraystretch}{1.06}
\begin{tabular*}{\linewidth}{@{\extracolsep{\fill}}lrrr@{}}
\toprule
Task & Demos & \shortstack{Total\\updates} & Trials\\
\midrule
Handover Box & 50 & 50k & 200\\
Shoes Table & 50 & 40k & 200\\
Handover Mic & 50 & 30k & 200\\
Scan Object & 50 & 60k & 200\\
\bottomrule
\end{tabular*}
\end{wraptable}

The transfer study uses ALOHA Handover Box and Shoes Table, together with
RoboTwin 2.0 Handover Mic and Scan Object \citep{robotwin2}.
Models are fine-tuned from pretrained $\pi_{0.5}$ policies using the native
flow-matching action objective. Table~\ref{tab:app_transfer_setup} summarizes
the demonstrations, total adaptation, and evaluation size.
The Box and Shoes Full models use Common-first initialization;
Mic and Scan enable Full interaction directly.
Each task uses fixed model weights for evaluation and interaction removal.
In particular, Scan's Full and removal results use the same model with
60k total adaptation updates, trained on 50 demonstrations.

Evaluation conditions fix the task initialization and instruction across
compared settings. Observation assignment is unchanged between Full and
its controls: each policy receives its private wrist observation and state,
as well as the shared external view where supplied by the task.
Independent retains these inputs while removing all representation exchange
during both learning and inference. All-off instead disables the three
paths in an already trained Full policy.
These interventions retain different learning histories even when their
inference-time information interfaces coincide.

\subsection{Reference Settings and Additional Tasks}
\label{app:published_references}
\label{app:handover_block}
\label{app:additional_tasks}

\begin{table}[!htbp]
\centering\small
\caption{Reference evaluation settings. Trial counts are per task.}
\label{tab:app_reference_settings}
\begin{tabular*}{\linewidth}{@{\extracolsep{\fill}}lllr@{}}
\toprule
Methods & Tasks & Source / setting & Trials\\
\midrule
RDT-1B, $\pi_0$, TwinVLA & Box, Shoes & \citet{twinvla}, Table 9 / Easy & 500\\
TwinVLA & Mic, Scan & \citet{twinvla}, Table 10 / Easy & 100\\
RDT-1B, $\pi_0$ & Mic, Scan & \citet{robotwin_leaderboard} / Single, clean2clean & 100\\
\bottomrule
\end{tabular*}
\end{table}

Exploratory $\pi_{0.5}$ Full policies also reach 31.5\% on Handover Block
and 23.0\% on Hanging Mug, each over 200 development episodes.
Handover Block is distinct from the ALOHA Handover Box task.
These additional outcomes indicate remaining difficulty in multi-stage manipulation.

\subsection{Additional Interaction Controls}
\label{app:functional_controls}
\label{app:recent_controls}
\label{app:diagnostic_completion}
\label{app:mic_channel_removals}
\label{app:scan_channel_removals}

Table~\ref{tab:cross_backbone} reports the individual-path removals and
All-off results. In additional 200-episode evaluations, a separately
trained $\pi_{0.5}$ Independent policy achieves 63.5\% on Handover Box.
Disabling both Private Prefix K/V and action interaction in the Full
checkpoint, while retaining Common interaction, yields 53.5\% on Handover
Box and 82.5\% on Shoes Table. This is a joint inference-time removal
without retraining.
The Independent Box model uses 40k total updates, compared with 50k for
Full, so the training comparison has unequal budgets.

\FloatBarrier
\section{Scaling VLA-Fabric to Larger Teams}
\label{app:multiarm_recipe}

\subsection[From Two to N Local VLA Agents]{From Two to $N$ Local VLA Agents}
\label{app:team_organization}

The interaction functions in Appendix~\ref{app:functional_interface} extend
to a team by expanding each receiver's peer set.
For the evaluated fully connected teams,
$\mathcal N(i)=\{1,\ldots,N\}\setminus\{i\}$.
Each added agent retains its complete policy, local proprioception,
action-generation state, and local action head.
Common aggregation includes the additional agent, and Private Prefix K/V
are concatenated in the team's fixed role order.
The action branch attends to each peer separately, then concatenates the
resulting contexts before the receiver-local projection:
\begin{equation}
 u_{ij}^{k,s}
 =\operatorname{Attn}\!\left(Q_{i,\mathrm{act}}^{k,s},
 K_{j,\mathrm{act}}^{k,s},V_{j,\mathrm{act}}^{k,s}\right),\quad
 r_i^{k,s}
 =\alpha_i^k W_i^k
 \operatorname{Concat}^{\mathrm{role}}_{j\in\mathcal N(i)}u_{ij}^{k,s}.
\label{eq:multiarm_peer_context}
\end{equation}
This extends Equation~\ref{eq:pi_action_interaction} while keeping both
the receiver's query and the attention normalization for each peer local.
The local observation and action dimensions remain unchanged as arms are
added, although peer context and role-dependent modules can grow.
Each task and team configuration is fine-tuned separately.

\subsection{Simulation Tasks and Setup}
\label{app:simulation_setup}
\label{app:frame3}
\label{app:frame_controls}

Table~\ref{tab:app_simulation_setup} specifies the team sizes and
completion criteria for the larger-team tasks. Stack3 uses RoboFactory;
Frame3, Frame4, and Arch4 use our MuJoCo task environments.
Agents receive local wrist images and proprioception, together with the
global scene view supplied as shared context.
Full and Independent use the same task data and observation assignment.
Independent disables Common, Private K/V, and action interaction throughout
fine-tuning and inference.

\begin{table}[!htbp]
\centering\small
\caption{Larger-team simulation settings. Each policy setting uses 200
evaluation trials; completion requires the full task outcome.}
\label{tab:app_simulation_setup}
\begin{tabular*}{\linewidth}{@{\extracolsep{\fill}}lrp{0.76\linewidth}@{}}
\toprule
Task & Arms & Success criterion\\
\midrule
Stack3 & 3 & Complete the ordered cube stack with all assigned arms released.\\
\addlinespace
Frame3 & 3 & Place the triangular frame over the peg after joint lifting
and transport; release and retreat with the frame stable.\\
\addlinespace
Frame4 & 4 & Place the quadrilateral frame over the peg after joint lifting
and transport; release and retreat with the frame stable.\\
\addlinespace
Arch4 & 4 & Position the supports and place the beam to form a stable
completed arch.\\
\bottomrule
\end{tabular*}
\end{table}

The total Full adaptation budgets are 20k updates for Stack3, 30k each
for Frame3 and Frame4, and 50k for Arch4.
The Stack3 and frame Independent controls match the corresponding total
budgets. Frame4 nevertheless compares different adaptation recipes, so its
training comparison is distinct from a fixed-weight path removal.
Evaluations share task initializations, although rendered observations can
differ across policy runs. Each setting uses 200 trials, and episodes that
reach the horizon without completion remain failures.
Arch4 combines a longer assembly sequence, randomized object displacement,
and inter-arm collision risk.

\subsection{Additional Scaling Evidence}
\label{app:simulation_results}

Beyond the final comparisons in Table~\ref{tab:larger_team_main}, a direct-Full
Stack3 policy reaches 87.5\% over 200 development episodes after 10k total
updates. Frame3 also uses direct-Full adaptation.
These examples show that direct-Full adaptation can also learn the
transferred interaction functions.

Additional SingleVLA-Fabric teams, based on the Eagle2 VLM, perform
Camera Alignment and Take Photo in RoboFactory~\citep{robofactory}.
Camera Alignment requires two arms to lift a camera while a third holds
an object in front of it. Take Photo adds a fourth arm that presses the
camera button while the other arms maintain the camera--object arrangement.
Full achieves 100.0\% on three-arm Camera Alignment and 96.5\% on four-arm
Take Photo, each over 200 evaluation episodes.
On the same 200 Take Photo conditions, All-off reduces success to 17.5\%
by disabling Common, Private K/V, and action interaction in the Full
checkpoint without retraining.
These results extend the larger-team evidence to the Eagle2 VLM-based
realization and to coordinated holding and actuation, complementing the
$\pi_{0.5}$ tasks in the main text.

\subsection{Physical-Robot Setup and Evaluation}
\label{app:physical_protocol}
\label{app:physical_results}

\paragraph{Hardware and observations.}
The platform uses AgileX PiPER robot arms, each with six actuated joints
and a gripper, and Intel RealSense D455 cameras.
Each policy receives its local wrist RGB image, joint positions, and gripper
state. One global RGB view provides shared scene context in both Full and
Independent settings.

\paragraph{Demonstrations and action interface.}
Each VLA generates local chunks of six
joint-position targets and one gripper command per arm.
Each task uses 50 physical demonstrations for task-specific $\pi_{0.5}$
fine-tuning.

\paragraph{Tasks and evaluation.}
Table~\ref{tab:app_physical_tasks} defines the four tasks. Each Full and
Independent setting uses 25 physical trials; Independent learns without
interaction, whereas Full retains all three functions.
Success percentages and coordination sequences appear in
Table~\ref{tab:physical_main} and Figure~\ref{fig:real_robot_sequences}.

\begin{table}[!htbp]
\centering\small
\caption{Physical task definitions and successful completion.}
\label{tab:app_physical_tasks}
\begin{tabular*}{\linewidth}{@{\extracolsep{\fill}}lrp{0.76\linewidth}@{}}
\toprule
Task & Arms & Role assignment and completion criterion\\
\midrule
Basket3 & 3 & Each arm loads an assigned vegetable; success requires all
vegetables to be placed in the shared basket.\\
\addlinespace
Relay3 & 3 & One arm positions the receiving target, while two arms transfer
the object; success requires handover and placement within the target.\\
\addlinespace
Relay4 & 4 & Two sender--receiver pairs relay vegetables to the basket;
both pairs must complete their handovers and basket placements.\\
\addlinespace
Frame4 & 4 & Four arms jointly lift, align, and lower the frame over the peg;
success requires stable placement after release.\\
\bottomrule
\end{tabular*}
\end{table}

\FloatBarrier

\end{document}